\pdfoutput=1

\documentclass[11pt]{article}

\usepackage{EMNLP2022}

\usepackage{times}
\usepackage{latexsym}

\usepackage[T1]{fontenc}

\usepackage[utf8]{inputenc}

\usepackage{microtype}

\usepackage{inconsolata}

\usepackage{booktabs}
\usepackage{multirow}
\usepackage{graphicx}
\usepackage{caption}
\usepackage{subcaption}
\usepackage{amsfonts}
\usepackage{algorithm}
\usepackage{algorithmic}
\usepackage{amssymb}
\usepackage{pifont}

\definecolor{deepblue}{rgb}{0,0,0.5}
\definecolor{officeblue}{RGB}{0,102,204}
\definecolor{deepred}{rgb}{0.6,0,0}
\definecolor{deepgreen}{rgb}{0,0.5,0}
\definecolor{mybrickred}{RGB}{182,50,28}
\definecolor{fillcolor}{RGB}{216,217,252}

\newcommand{\sptk}[1]{\texttt{[#1]}}
\renewcommand{\algorithmiccomment}[1]{\bgroup\hfill $\triangleright$ ~#1\egroup}

\usepackage{amsmath,amsfonts,bm}

\def\eqref#1{equation~\ref{#1}}

\def\1{\bm{1}}

\def\vt{{\bm{t}}}

\def\vv{{\bm{v}}}
\def\vw{{\bm{w}}}

\def\vz{{\bm{z}}}

\def\mA{{\bm{A}}}

\def\mH{{\bm{H}}}

\def\mK{{\bm{K}}}

\def\mQ{{\bm{Q}}}

\def\mT{{\bm{T}}}

\def\mV{{\bm{V}}}
\def\mW{{\bm{W}}}

\DeclareMathAlphabet{\mathsfit}{\encodingdefault}{\sfdefault}{m}{sl}
\SetMathAlphabet{\mathsfit}{bold}{\encodingdefault}{\sfdefault}{bx}{n}

\newcommand{\R}{\mathbb{R}}

\newcommand{\softmax}{\mathrm{softmax}}

\newcommand\our{\textsc{DiDE}}
\newcommand{\cmark}{\textcolor{green}{\ding{51}}}
\newcommand{\xmark}{{\color{red}\ding{55}}}

\title{Distilled Dual-Encoder Model for Vision-Language Understanding}

\author{
    Zekun Wang$^\dag$\thanks{~~Contribution during internship at Microsoft Research.},~~Wenhui Wang$^\ddag$,~~Haichao Zhu$^\dag$,~~Ming Liu$^{\dag\natural}$,~~Bing Qin$^{\dag\natural}$,~~Furu Wei$^\ddag$ \\
$^\dag$Harbin Institute of Technology, Harbin, China \\
$^\ddag$Microsoft Research, Beijing, China \\
$^\natural$Peng Cheng Laboratory, Shenzhen, China \\
\texttt{\{zkwang,hczhu,mliu,qinb\}@ir.hit.edu.cn} \\
\texttt{\{wenwan,fuwei\}@microsoft.com}
}

\begin{document}
\maketitle

\begin{abstract}
On \textbf{v}ision-\textbf{l}anguage \textbf{u}nderstanding (VLU) tasks, fusion-encoder vision-language models achieve superior results but sacrifice efficiency because of the simultaneous encoding of images and text.
On the contrary, the dual-encoder model that separately encodes images and text has the advantage in efficiency, while failing on VLU tasks due to the lack of deep cross-modal interactions.
To get the best of both worlds, we propose \our{}\footnote{Our code and models will be publicly available at \url{https://github.com/kugwzk/DiDE}.}, a framework that distills the knowledge of the \textbf{fusion-encoder teacher} model into the \textbf{dual-encoder student} model. 
Since the cross-modal interaction is the key to the superior performance of teacher model but is absent in the student model, we encourage the student not only to mimic the predictions of teacher, but also to calculate the cross-modal attention distributions and align with the teacher.
Experimental results demonstrate that \our{} is competitive with the fusion-encoder teacher model in performance (only a $1\%$ drop) while enjoying $4\times$ faster inference.
Further analyses reveal that the proposed cross-modal attention distillation is crucial to the success of our framework.

\end{abstract}
\section{Introduction}
\begin{figure}[t]
    \begin{center}
    \includegraphics[trim={0 14 0 7.4}, clip, width=\linewidth]{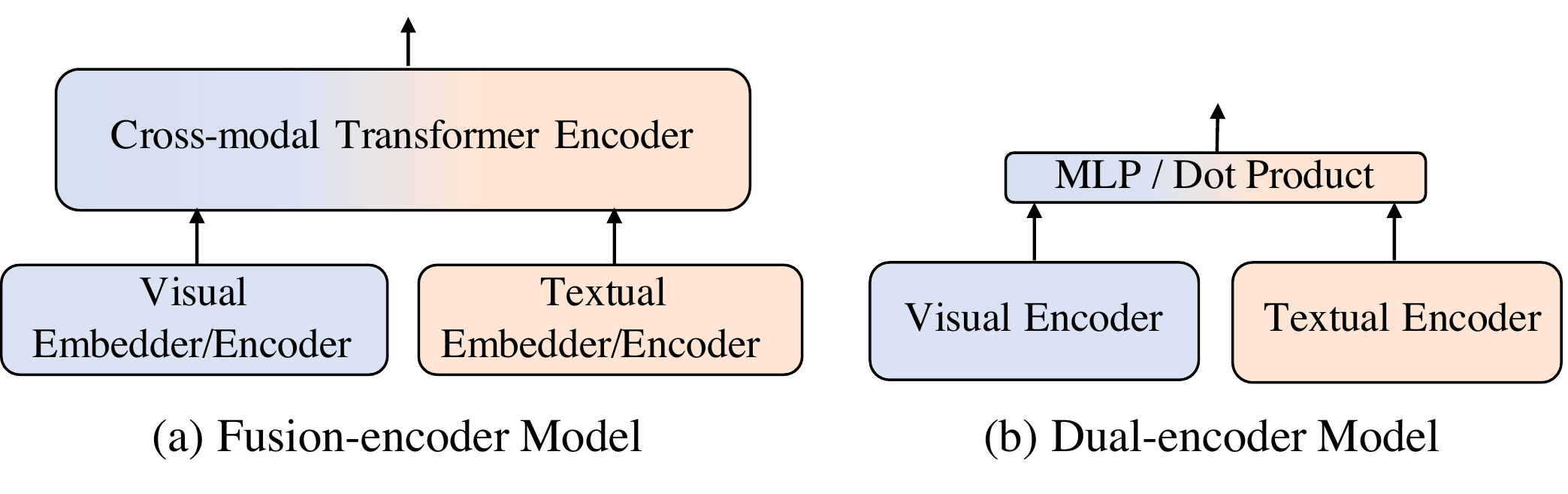}
    \caption{ 
        Illustration of two architectures of vision-language models.
        (a) Fusion-encoder models simultaneously encode visual and textual inputs via modal-specific embedders/encoders and employ a cross-modal Transformer encoder to fuse representations.
        (b) Dual-encoder models encode images/text separately and adopt an extreme lightweight module (\textit{e.g.}, MLP) for cross-modal interactions.
    }
    \label{fig:arch}
    \end{center}
\end{figure}

Vision-language understanding (VLU) tasks (\textit{e.g.}, visual reasoning~\cite{NLVR2}, visual entailment~\cite{snli_ve}, visual question answering~\cite{vqav2}) require the model to understand the cross-modal interactions between images and text.
Various fusion-encoder vision-language pretrained models~\citep{lxmert,uniter,vinvl,vilt,meter,Flamingo} are proposed for VLU tasks.
As shown in Figure~\ref{fig:arch}(a), these models employ a Transformer~\citep{transformer} network as a cross-modal encoder to capture interactions between different modalities.
Despite the remarkable performance, the heavy cross-modal encoder remains an efficiency bottleneck due to the simultaneous encoding of images and text, limiting the application in practical scenarios with massive images or text.
Therefore, it is crucial to find an approach to accelerate inference for VLU.

We turn to explore the dual-encoder vision-language model (shown in Figure~\ref{fig:arch}(b)), which encodes images and text separately and then applies an extreme lightweight shallow module to model the interactions between modalities.
The disentangled encoding paradigm enables off-line computing and caching visual or textual representations on demand, significantly lowering runtime latency.
However, the shallow module is insufficient to handle complex VLU tasks, resulting in previous models~\cite{clip,align} falling far behind fusion-encoder models~\citep{vilt}.
Can dual-encoder models obtain performance comparable to fusion-encoder models while preserving efficiency?
In this work, we propose \our{} (a knowledge \textbf{Di}stillation framework for \textbf{D}ual-\textbf{E}ncoder models), where the dual-encoder model (student) is supervised by the fusion encoder models (teacher), as shown in Figure~\ref{fig:model}.
Although soft label distillation~\cite{hinton_kd} is widely applied,
our key observation is that \textit{cross-modal attention distributions}\footnote{visual-to-textual (blue) and textual-to-visual (orange) attention in Figure~\ref{fig:model}.} are absent in the student, resulting in the inability to model complex cross-modal interactions.
Thus, only distilling soft labels is not enough for the student to mimic the interactions of teacher deeply.

Considering that cross-modal interaction is critical for VLU, we introduce a plug-and-play objective \textit{cross-modal attention distillation}, as fine-grained supervision to help the student better learn cross-modal interactions.
Specifically, besides the soft label distillation, we compute the cross-modal attention of the student model and align it with the distribution in the teacher model during training.
The training of \our{} consists of two-stage distillations.
In the pre-training stage, the student learns a general initialization with distillation.
In the fine-tuning stage, distillation helps the student learn more task-specific knowledge.
Experimental results demonstrate that \our{} performs competitively with the fusion-encoder teacher model in various VLU tasks (retaining $96.9\%$ to $99.9\%$ performance) while having a $4\times$ inference speedup.
Further analysis indicates that the proposed cross-modal attention distillation yields significant gains compared to distilling only with soft labels or other latent features of the teacher.
Beyond VLU, \our{} also shows effectiveness in image-text retrieval.

Our contributions are summarized as follows:
\begin{itemize}
\item We propose \our{}, a knowledge distillation framework for the dual-encoder model to learn better cross-modal interactions of vision-language understanding from the fusion-encoder model.
\item Our approach is plug-and-play with different vision-language tasks and can be applied on different model architectures.
\item Experimental results show that our distilled model performs competitively with the teacher model and has a significant speedup. Further analysis indicates that our proposed cross-modal attention distillation is the key to success.
\end{itemize}
\section{Related Work}
\subsection{Vision-Language Pre-Training}
Language and vision pre-training advance the state of the art in downstream natural language processing tasks~\cite{gpt,bert,unilm,roberta,unilm2,bart,t5,xlmr,xlme} and computer vision tasks~\cite{vit,deit,beit}.
Vision-Language pre-training~\cite{vilbert,vlbert, villa, unimo, vlmo, simvlm} has been shown to prevail in learning cross-modal representations.
The model architectures fall into two lines: \textit{fusion-encoder} and \textit{dual-encoder} models.
Fusion-encoder models jointly encode image-text pairs and employ a multi-layer cross-modal Transformer encoder to fuse the visual and textual representations.
Previous models~\cite{visualbert,lxmert,uniter, oscar, vinvl} extract visual features through a pre-trained object detector (\textit{e.g.}, Faster R-CNN~\cite{faster_rcnn}), which requires high-resolution input images and brings more computation costs.
\citet{pixelbert, albef, meter} directly take image pixels or patches as input and encode visual features by CNN or Vision Transformer~\cite{vit}.
ViLT~\citep{vilt} directly applies a shared Transformer for joint encoding of image patches and textual token embeddings, achieving competitive performance with less overhead.
The models exhibit a strong ability to model complex cross-modal interactions and achieve superior results on VLU tasks.
However, the models rely on a cross-modal Transformer encoder to fuse visual and textual features simultaneously across layers, demanding a heavy computation budget and leading to a low inference speed.

On the contrary, dual-encoder models~\cite{clip,align, lightningdot} encode images and text separately and take an MLP or dot product to model the interactions between the modalities.
These models have the advantage of computational efficiency.
The attention mechanism is computed only within tokens of the same modality.
Moreover, thanks to the independent encoders, the visual or textual representations can be precomputed and cached off-line in the practical scenarios.
However, the shallow module is not enough to handle complex cross-modal interactions, causing significant performance degradation on VLU tasks~\citep{vilt, DBLP:journals/tacl/HendricksMSAN21}.
To get the best of both worlds, we preserve the inference efficiency of dual-encoder model while achieving promising results on VLU by knowledge distillation.
\subsection{Knowledge Distillation}
Knowledge distillation (KD;~\citet{hinton_kd} aims to improve a student model by transferring knowledge from a teacher model.
Transformer distillation is widely used in various domains~\cite{tinybert, minilmv1, deit, kd_on_vlm}.
In this work, we focus on distillation under the cross-architecture setting~\cite{cross_model_distill}, where the architectures of the teacher and the student are different.
\citet{deformer} decomposes the early layers of the Transformer and adopts a complete Transformer to guide the training for reading comprehension.
In image-text retrieval, \citet{think_fast_slow} proposes distilling soft labels from a cross-attention model to a dual-encoder model with the reranking mechanism.
But in VLU tasks, their method does not work, whereas our proposed cross-modal attention distillation is critical for success.
\section{Method}

\begin{figure*}[!t]
    \begin{center}
    \includegraphics[trim={0 7.5 0 6.5}, clip, width=1.0\textwidth]{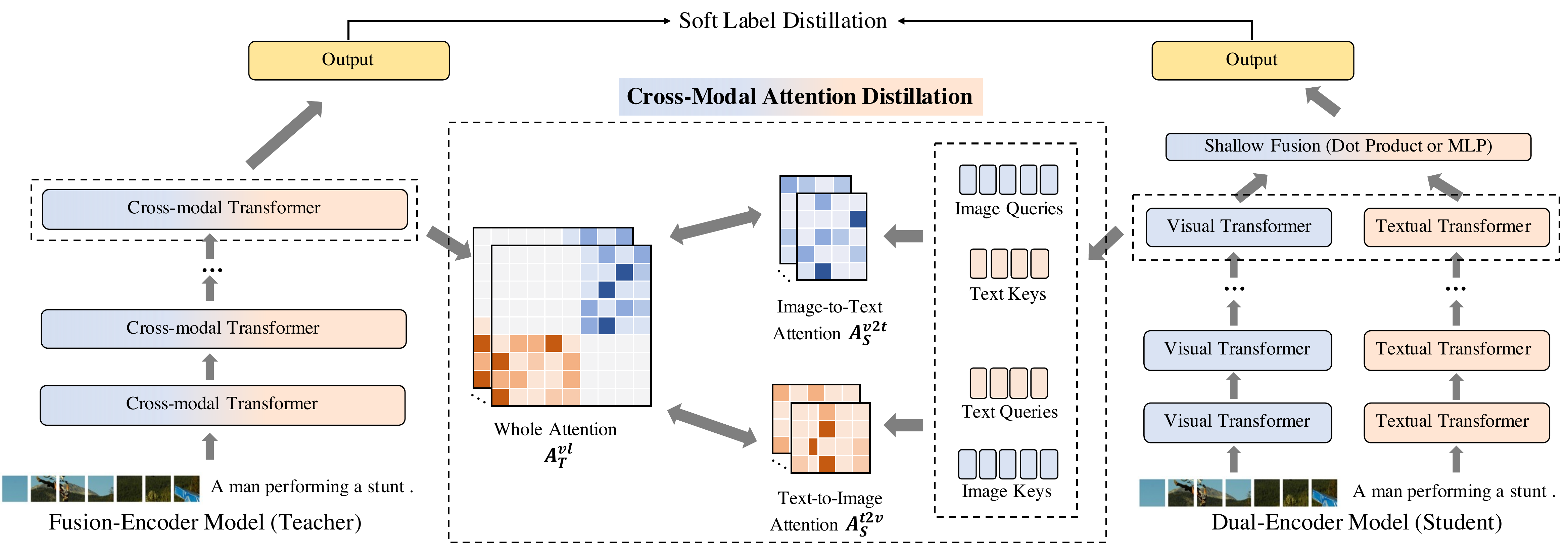}
    \caption{ 
        Overview of our framework \our{}, best viewed in color.
        Besides soft labels, we introduce cross-modal attention distillation to guide the model training.
        The visual-to-textual attention $\mA^{\textit{v2t}}$ (blue) and the textual-to-visual $\mA^{\textit{t2v}}$ (orange) of the dual-encoder model (student) are aligned to the fusion-encoder model (teacher).
        Other part of the attention distributions (grey) are omitted.
    }
    \vspace{-5mm}
    \label{fig:model}
    \end{center}
\end{figure*}

Figure~\ref{fig:model} gives an overview of our \our{} framework, a knowledge distillation approach for the dual-encoder model.
\subsection{Model Overview}

\paragraph{Input Representations} 
We slice the input image $\vv \in \R^{H \times W \times C}$ into patches $\vv^{\textit{p}} \in \R^{N \times (P^2 C)}$, where $N={HW}/{P^2}$ is the number of patches, $(H, W)$ is the resolution of the input image, $(P, P)$ is the resolution of each patch, and $C$ is the number of channels.
The input text $\vt$ is tokenized into a sequence of $M$ tokens.
We prepend the special tokens \sptk{I\_CLS} and \sptk{T\_CLS} to the sequence of image patches and text tokens, respectively.
We linearly project image patches $\vv^{\textit{p}}$ to obtain patch embeddings, and the final visual input embeddings $\mH^{\textit{v}}_{0} \in \R^{(N+1) \times D}$ are computed via:
\begin{gather*}
    \mH_0^{\textit{v}} = [ \vv_{\sptk{I\_CLS}} , \mV \vv^{\textit{p}}_{1} , \dots , \mV \vv^{\textit{p}}_{N} ] + \mV_{\textit{pos}} + \mV_{\textit{type}}
\end{gather*}
where $\mV \in \R^{(P^2 C) \times D}$ is linear projection, $\mV_{\textit{pos}} \in \R^{(N+1) \times D}$ is 1D positional embedding, $\mV_{\textit{type}} \in \R^{D}$ is visual type embedding.
The textual input embeddings $\mH^{\textit{t}}_{0} \in \R^{(M+1) \times D}$ are obtained by summing word embeddings $\mW$, the textual position embedding $\mT_{\textit{pos}}$ and textual type embedding $\mT_{\textit{type}}$:
\begin{gather*}
   \mH_0^{\textit{t}} = [ \vw_{\sptk{T\_CLS}} , \vw_{1} , \dots , \vw_{M} ] + \mT_{\textit{pos}} + \mT_{\textit{type}} 
\end{gather*}
We take $\mH_0^{\textit{v}}, \mH_0^{\textit{t}}$ as visual and textual inputs for the teacher and student models.

\paragraph{Fusion-Encoder Model (Teacher)} 
concatenates the input representations $\mH^{\textit{v}}_{0}$ and $\mH^{\textit{t}}_{0}$ as $\mH^{\textit{vl}}_{0} = [ \mH^{\textit{v}}_{0}; \mH^{\textit{t}}_{0} ]$, and feeds into a $L$-layer cross-modal Transformer encoder to obtain contextual representations $\mH_{L}^{\textit{vl}}$.
The cross-modal Transformer encoder fuses representations of different modalities via the multi-head attention mechanism.
Specifically, for each head $a$, the whole attention distribution $\mA_{a}^{\textit{vl}}$ is computed via:
\begin{gather*}
    \mA_{a}^{\textit{vl}} = \softmax( \frac{ \mQ_{a}^{\textit{vl}}\mK_{a}^{\textit{vl} \intercal}}{ \sqrt{d_k}})
    \label{eq:attention}
\end{gather*}
where queries $\mQ^{\textit{vl}}_{a}$ and keys $\mK_{a}^{\textit{vl}}$ are obtained by linearly projecting the hidden states using parameters $\mW^{\textit{Q}}_{l,a}, \mW^{\textit{K}}_{l,a} \in \R^{D \times d_{k}}$, respectively.
$d_{k}$ is the size of the attention head.
The output vectors of \sptk{I\_CLS} and \sptk{T\_CLS} are fed into the task-specific layer to obtain predictions.

\paragraph{Dual-Encoder Model (Student)}
encodes $\mH_{0}^{\textit{v}}$ and $\mH_{0}^{\textit{t}}$ are separately via visual and textual Transformer encoders: $\mH_{l}^{\textit{v}}$ and $\mH_{l}^{\textit{t}}$.
The output vectors of \sptk{I\_CLS} and \sptk{T\_CLS} are used as final representations of the images and text.
Then, a shallow module $f$ is applied to fuse the two representations.
For vision-language understanding, we adopt an MLP as the module $f$.
For image-text retrieval, we use the dot product function to obtain similarity scores of image-text pairs.

\subsection{Distillation Objectives}

\paragraph{Cross-Modal Attention Distillation}
The Transformer captures fine-grained interactions between tokens, mainly benefiting from the multi-head attention mechanism~\citep{attr}.
In the dual-encoder model, tokens only compute attention with those within the same modality,  named \textit{uni-modal attention}: visual-to-visual $\mA^{\textit{v2v}} \in \R^{N \times N}$ and textual-to-textual $\mA^{\textit{t2t}} \in \R^{M \times M}$.
As shown in Figure~\ref{fig:model}, compared to the whole attention $\mA^{\textit{vl}} \in \R^{(N+M)\times (N+M)}$ of the fusion-encoder model, we observe that the attention of computing intermodality tokens (named \textit{cross-modal attention}\footnote{We name the uni-modal attention and cross-modal attention from the perspective of dual-encoder models.}) is absent in the student, including visual-to-textual $\mA^{\textit{v2t}} \in \R^{N \times M}$ and textual-to-visual $\mA^{\textit{t2v}}\in\R^{M \times N}$.
Thus, the student lacks the ability to capture cross-modal interactions, which is critical for vision-language tasks.

Taking into account the weakness of the dual-encoder model, we propose the \textit{cross-modal attention distillation} objective.
Specifically, during training, the student calculates the cross-modal part of the attention distribution and mimics it with the teacher.
The cross-modal attention distributions of the student $\mA^{\textit{v2t}}_{\textit{S}}, \mA^{\textit{t2v}}_{\textit{S}}$ are computed as follows:
\begin{gather*}
    \mA^{\textit{v2t}}_{\textit{S}} = \softmax( \frac{\mQ^{\textit{v}}_{\textit{S}} \mK_{\textit{S}}^{\textit{t}\intercal}}{ \sqrt{d_k}}) \\
    \mA^{\textit{t2v}}_{\textit{S}} = \softmax( \frac{\mQ^{\textit{t}}_{\textit{S}} \mK_{\textit{S}}^{\textit{v} \intercal{}}}{ \sqrt{d_k}}) 
    \label{eq:cross_modal_attention}
\end{gather*}
where $\mQ^{\textit{v}}_{\textit{S}}, \mK^{\textit{v}}_{\textit{S}}$ are visual queries and keys of the attention module in the student.
$\mQ^{\textit{t}}_{\textit{S}}, \mK^{\textit{t}}_{\textit{S}}$ are queries and keys for textual features.
To better align the student, we calculate the cross-modal attention $\mA^{\textit{v2t}}_{\textit{T}}, \mA^{\textit{t2v}}_{\textit{T}}$ of the teacher in a way similar to above, instead of directly splitting the whole attention $\mA^{\textit{vl}}_{\textit{T}}$.
We use the cross-modal attention distillation loss to minimize the KL-divergence of the cross-modal attention distribution:
\begin{gather*}
\small
    \mathcal{L}_{\textit{CA}} = D_{\textit{KL}}(\mA^{\textit{v2t}}_{\textit{S}} \parallel \mA^{\textit{v2t}}_{\textit{T}}) + D_{\textit{KL}}(\mA^{\textit{t2v}}_{\textit{S}} \parallel \mA^{\textit{t2v}}_{\textit{T}})
    \label{eq:cssattn_loss}
\end{gather*}
We empirically find that only distilling between the last layer of teacher and student is more effective (detailed in Section~\ref{section:discussion}).

\paragraph{Soft Label Distillation}
In addition to cross-modal attention distillation, we also apply the soft label distillation loss to align the predictions between the teacher and the student:
\begin{gather*}
    \mathcal{L}_{\textit{SL}} = D_{\textit{KL}}(\vz_{\textit{S}} \parallel \vz_{\textit{T}})
    \label{eq:soft_label}
\end{gather*}
where $\vz_{\textit{S}}, \vz_{\textit{T}}$ are the output logits of the student and the teacher, respectively.

\subsection{Two-Stage Distillation Training}
\label{section:framework}
\our{} applies the distillation objectives via the prevalent two-stage training paradigm: pre-training and then fine-tuning.

\subsubsection{Pre-Training Distillation}

We consider three typical pre-training tasks: cross-modal contrastive learning, image-text matching, and masked language modeling.
\paragraph{Cross-Modal Contrastive Learning (CMC)}
We introduce an InfoNCE contrastive loss~\citep{infonce} with in-batch negative sampling to optimize the shared space of visual and textual representations.
Specifically, the image-to-text contrastive loss is computed as:
\begin{gather*}
    \mathcal{L}_{\textit{NCE}}^{\textit{i2t}} = - \sum_{i=1}^{N} \log \frac{\exp(\text{sim}(\textit{I}_{i},\textit{T}_{i}) / \tau )}{ \sum_{j=1}^{N} \exp(\text{sim}(\textit{I}_{i},\textit{T}_{j})/ \tau)}
    \label{eq:i2t_loss}
\end{gather*}
where $\tau$ is a trainable temperature parameter, $I_{i}$ and $T_{i}$ are representations of the $i$-th image-text pair in the batch.
We use the dot product as the $\text{sim}(\cdot,\cdot)$ function.
Similarly, the text-to-image contrastive loss is computed as follows:
\begin{gather*}
    \mathcal{L}_{\textit{NCE}}^{\textit{t2i}} = - \sum_{i=1}^{N} \log \frac{\exp(\text{sim}(\textit{T}_{i},\textit{I}_{i}) / \tau )}{ \sum_{j=1}^{N} \exp(\text{sim}(\textit{T}_{i},\textit{I}_{j})/ \tau)}
    \label{eq:t2i_loss}
\end{gather*}
For the soft label distillation, the fusion-encoder model requires joint encoding of each image-text pair, which results in quadratic time complexity to obtain outputs.
To reduce the training computation, we omit the soft label loss while only applying the cross-modal attention distillation on $N$ matched pairs with gold labels:
\begin{gather}
    \mathcal{L}^{\textit{CMC}} = \mathcal{L}_{\textit{NCE}}^{\textit{i2t}} + \mathcal{L}_{\textit{NCE}}^{\textit{t2i}} + \mathcal{L}_{\textit{CA}}^{\textit{CMC}}
    \label{eq:cmc}
\end{gather}
\paragraph{Image-Text Matching (ITM)}
The goal of image-text matching is to predict whether the input image and text are matched.
We employ cross-modal attention distillation loss over the input pairs and soft-label loss for training:
\begin{gather*}
\mathcal{L}^{\textit{ITM}} = \mathcal{L}_{\textit{CA}}^{\textit{ITM}} + \mathcal{L}_{\textit{SL}}^{\textit{ITM}}
\end{gather*}
\paragraph{Masked Language Modeling (MLM)}
Masked language modeling aims to recover the masked tokens from the other unmasked tokens.
Similarly to ITM, the student needs to mimic both cross-modal attention distributions and soft labels from the teacher: 
\begin{gather*}
    \mathcal{L}^{\textit{MLM}} =  \mathcal{L}_{\textit{CA}}^{\textit{MLM}} + \mathcal{L}_{\textit{SL}}^{\textit{MLM}}
\end{gather*}

\subsubsection{Fine-Tuning Distillation}
\paragraph{Vision-Language Understanding}
For VLU tasks, the student is fine-tuned with cross-modal attention and soft label distillation objectives:
\begin{gather*}
    \mathcal{L}^{\textit{VLU}} =  \mathcal{L}_{\textit{CA}}^{\textit{VLU}} + \mathcal{L}_{\textit{SL}}^{\textit{VLU}}
\end{gather*}
\paragraph{Image-Text Retrieval}
For image-text retrieval, the student is fine-tuned on the image-text retrieval task with the same objective as the CMC task (Equation~\ref{eq:cmc}).
\section{Experiments}
\begin{table}[!t]
    \small
	\centering	
    \scalebox{1.0}{
	\begin{tabular}	{l | c c c c }
	\toprule	 	
	 \textbf{Datasets} & \textbf{NLVR2} & \textbf{SNLI-VE} & \textbf{VQA} & \textbf{Flickr30K} \\
	 \midrule
     \#Images & 119K & 31K & 204K & 32K \\
	 \#Texts & 100K & 565K & 1.1M & 160K \\
	\bottomrule
	\end{tabular}
 	}
\caption{
		Statistics of downstream VL datasets.
	}
	\label{table:datasets}
\end{table}

\subsection{Datasets}
We use four commonly used datasets for pre-training: COCO~\cite{coco_dataset}, Conceptual Captions~\cite{cc_dataset}, SBU Captions~\cite{sbu_dataset} and Visual Genome~\cite{vg_dataset}, with in total $4$M images.
We experiment on three vision-language understanding datasets and one image-text retrieval fine-tuning dataset.
Table~\ref{table:datasets} shows the statistics of datasets.

\paragraph{Natural Language for Visual Reasoning}
The NLVR2 dataset~\cite{NLVR2} is a visual reasoning task that aims to determine whether a textual statement describes a pair of images.
Following previous work~\cite{uniter, vilt}, we construct two pairs of image-text, each consisting of the image and a textual statement.
The representations of the two pairs are fed into a classifier layer to obtain the final prediction.

\paragraph{Visual Entailment}
The SNLI-VE~\cite{snli_ve} is a three-way classification dataset, aiming to predict the relationship between an image and a text hypothesis: \textit{entailment}, \textit{natural}, and \textit{contradiction}.

\paragraph{Visual Question Answering}
The task requires the model to answer questions based on the input image.
We evaluate on the widely used VQAv2~\cite{vqav2} dataset.
Following~\citet{DBLP:conf/cvpr/00010BT0GZ18}, we formulate the problem as a classification task with 3,129 answer candidates.

\paragraph{Image-Text Retrieval}
The task consists of two subtasks: image retrieval and text retrieval.
We experiment on the Flickr30K~\citep{Flickr} with the standard split~\citep{Karpathy_split}.

\begin{table*}[!t]
\centering
\small

\begin{tabular}{l|ccccc|c}
\toprule
\multirow{2}{*}{\textbf{Models}} & \multicolumn{2}{c}{\textbf{NLVR2}} & \multicolumn{2}{c}{\textbf{SNLI-VE}} & \textbf{VQA} & \textbf{Inference} \\
& dev & test-P & val & test & test-dev & \textbf{Speedup} \\
\midrule
\multicolumn{7}{l}{~~\textit{Fusion-encoder models without using object region features}} \\
PixelBERT-R50 & 71.7 & 72.4 & - & - & 71.4 & 0.2$\times$\\ 
ViLT (Teacher) & 75.7 & 76.1 & 76.6 & 76.4 & 71.3 & 1.0$\times$\\
\midrule
\multicolumn{7}{l}{~~\textit{Dual-encoder models without using object region features}} \\
CLIP$^{\dag}$ & 50.9 & 51.1 & 68.4 & 68.6 & 50.2 & 4.1$\times$ \\
SLIP$^{\dag}$ & 50.9 & 51.1 & 70.9 & 71.0 & 55.9 & 4.1$\times$ \\
DeCLIP$^{\dag}$ & 50.9 & 51.1 & 69.9 & 70.2 & 59.6 & 4.1$\times$ \\
\our{}(Ours) & \textbf{75.3} & \textbf{75.6} & \textbf{76.5} & \textbf{76.3} & \textbf{69.2} & 4.0$\times$ \\
\midrule
\midrule
\multicolumn{7}{l}{~~\textit{Using object region features from the object detector}} \\
VisualBERT & 67.4 & 67.0 & - & - & 70.8 & $\ll 1.0\times$ \\
LXMERT & 74.9 & 74.5 & - & - & 72.4 & $\ll 1.0\times$\\
UNITER-Base & 75.9 & 75.8 & 78.6 & 78.3 & 72.7 & $\ll 1.0\times$\\
\bottomrule
\end{tabular}

\caption{
		Results on vision-language understanding tasks.
		The results are averaged over $4$ runs.
		We report vqa-score on VQA, accuracy for NLVR2 and SNLI-VE.
		$\dag$ is our reimplementation of fine-tuning, which is the same as \our{}.
		We evaluate the inference speed of dual-encoder models and ViLT on the NLVR2 dataset with the same hyper-parameters.
		The inference speedup of other models is taken from~\citet{vilt}.
}
\label{table:main_vlu}
\end{table*}

\subsection{Implementation Details}
In the main experiments, we use ViLT~\citep{vilt} as our teacher due to its simplicity and effective performance.
The visual and textual Transformers of the student model \our{} consist of $12$-layer blocks with $768$ hidden size and $12$ attention heads.
The intermediate size of feed-forward networks is $3072$.
Following~\citet{vilt}, the images are resized to $384 \times 640$ resolution and the patch size is $32 \times 32$.
The maximum length of the text sequence is set to $40$.
We optimize \our{} with Adam~\citep{adam} using a batch size of $1024$ for a total of $200$K steps on 16 Nvidia V100 GPU cards.
Note that our computation is less than the previous dual-encoder and fusion-encoder models.
Refer to Appendix~\ref{sec:app:hyperparam} for more details.

For the inference stage, we cache visual representations\footnote{Acturally, we can cache the visual or textual features according to the situation to improve efficiency.} for two reasons: 
(1) the averaged length of the visual tokens is longer than the textual tokens (240 \textit{vs.} 40).
(2) As shown in Table~\ref{table:datasets}, an input image is combined with multiple text sentences.
We reuse the cached visual representations with different text inputs to lower the inference latency.

\subsection{Results}
\label{section:results}

\begin{table}[!t]
    \centering
    \small
    \resizebox{\linewidth}{!}{
	\begin{tabular}	{l | c  c  c }
	\toprule	 	
	 \textbf{Models} & \textbf{NLVR2} & \textbf{SNLI-VE} & \textbf{VQA} \\
	  \midrule
	  \multicolumn{4}{l}{~~\textit{Online inference time}} \\
      ViLT (Teacher) & 150.3s & 189.4s & 1103.9s \\
      \our{} & 37.6s (4.0$\times$) & 49.7s (3.8$\times$) & 299.6s (3.7$\times$) \\
      \midrule
      \multicolumn{4}{l}{~~\textit{Offline cache time}} \\
        \our{} & 42.5s & 10.6s & 307.2s \\ 
	  \bottomrule
	\end{tabular}
	}
    \caption{
        Averaged inference and cache time (in seconds) of our model and teacher model ViLT on three VLU datasets.
        The inference time and cache time are evaluated on a P100 GPU with a batch size of 32.
    }
    \label{table:infer_time}
\end{table}
\begin{table*}[!t]
    \small
	\centering
	\begin{tabular}	{l|cccccc|c}
	\toprule	 	
	 \multirow{2}{*}{\textbf{Models}} & \multicolumn{3}{c}{\textbf{Image Retrieval}} & \multicolumn{3}{c}{\textbf{Text Retrieval}}  & \textbf{Inference}\\
	  & R@1 & R@5 & R@10 & R@1 & R@5 & R@10 & \textbf{Speedup}\\
	  \midrule
	   ViLT (Teacher) & 64.4 & 88.7 & 93.8 & \textbf{83.5} & \textbf{96.7} & 98.6 & 40398s \\
      \our{} & \textbf{68.2} & \textbf{89.8} & \textbf{94.2} & 83.2 & \textbf{96.7} & \textbf{98.8} & 16.1s (2509.2$\times$)\\
	  ~~$-$Cross-modal attention& 66.6 & 89.2 & 93.4 & 81.6 & 95.6 & 98.4 & -\\
	\bottomrule
	\end{tabular}
	\caption{
		Retrieval results on the Flickr30K dataset.
		``$-$Cross-modal attention'' is ablation trained without cross-modal attention distillation.
		The inference speed of our model and ViLT is evaluated under the same setup.
	}
	\label{table:retrieval_results}
	
\end{table*}		

\paragraph{Vision-Language Understanding}
We compare \our{} with three types of vision-language pretrained models: 
(1) Dual-encoder models. CLIP~\citep{clip} is pre-trained with image-text contrastive loss on $400$M image-text pairs, significantly more than our pre-training data.
SLIP~\citep{SLIP} and DeCLIP~\citep{declip} are improvements of CLIP.
SLIP introduces self-supervised contrastive loss to CLIP.
DeCLIP further leverages widespread supervision among the image-text pairs.
For a fair comparison, we fine-tune them with the same range of hyperparameters as \our{}.
(2) Fusion-encoder models without the object detector, such as our teacher ViLT~\citep{vilt}. 
(3) Fusion-encoder models with the object detector, such as UNITER~\citep{uniter}, that need a pretrained object detector to extract image region features, bringing more computational overhead.

Table~\ref{table:main_vlu} presents the performance of the VLU tasks.
The results are averaged over four random seeds. 
\our{} achieves competitive performance compared to the ViLT teacher (retaining $99.3\%$ in NLVR2, $99.9\%$ in SNLI-VE, and $96.9\%$ in VQA) while enjoying a $4$ times speedup.
\our{} significantly outperforms previous dual-encoder baselines (CLIP and its variants) by a large margin.
It is worth mentioning that dual-encoder baselines only achieve chance-level accuracy on the complex visual reasoning dataset NLVR2, while our \our{} obtains promising results (from $50.9$ to $75.3$ points).
To the best of our knowledge, it is the first demonstration that the dual-encoder model can obtain promising performance on the NLVR2 dataset.
Compared to other fusion-encoder models (with or without the pretrained object detector), \our{} also obtains comparable or even better results in VLU tasks.
This indicates that our approach can achieve a better efficiency-performance trade-off of the VLU.

\paragraph{Inference Speed}
We evaluate the latency of the student \our{} and the teacher ViLT in the batch inference setting, which is more favorable in low-latency scenarios~\cite{why_batch_infer}.
For \our{}, we pre-compute visual representations offline and cache them.
Table~\ref{table:infer_time} shows the averaged inference time measured on the test split of the datasets.
On all tasks, we obtain nearly $4$ times online inference speedup.
Even considering the offline cache time, \our{} is still faster than ViLT ($1.9\times$\textasciitilde$3.0\times$). 
Meanwhile, the storage cost is much cheaper than the cost of computing on GPUs~\citep{deformer}.
Thus, the dual-encoder model is more practical for production environments with massive inputs.
\begin{table*}[!t]
\centering
\small
\scalebox{0.95}{
    \begin{tabular}{l|cc|cc|cccccc}
    \toprule
    \multirow{2}{*}{\textbf{Models}} &
    \multicolumn{2}{c}{\textbf{Pre-training}} &
    \multicolumn{2}{c}{\textbf{Fine-tuning}} & \multicolumn{2}{c}{\textbf{NLVR2}} & \multicolumn{2}{c}{\textbf{SNLI-VE}} & \textbf{VQA} & \multirow{2}{*}{\textbf{Avg $\Delta$}} \\
    & STD & KD & STD & KD & dev & test-P & val & test & test-dev \\
    \midrule
    \our{} & \xmark & \cmark & \xmark & \cmark & \textbf{75.56} & \textbf{75.26} & \textbf{76.53} & \textbf{76.33} & \textbf{69.05} & -\\
    \midrule
    \multirow{5}{*}{Ablations} & - & - & \cmark & \xmark & 50.85 & 51.07 & 72.10 & 71.83 & 64.94 & -12.40 \\ 
    & - & - & \xmark & \cmark & 67.63 & 68.14 & 75.37 & 74.95 & 67.06 & -3.94 \\
    & \cmark & \xmark & \cmark & \xmark & 69.77 & 70.71 & 75.26 & 74.99 & 66.25 & -3.15 \\
    &  \cmark & \xmark & \xmark & \cmark & 73.06 & 74.11 & 76.11 & 75.87 & 67.10 & -1.30\\
    &  \xmark & \cmark & \cmark & \xmark & 70.98 & 71.40 & 75.30 & 75.21 & 66.84 & -2.60 \\
    
    \bottomrule
    \end{tabular}
}
\caption{
		Ablation results on vision-language understanding tasks.
        ``STD'' denotes training with original ground truth labels.
        ``KD'' denotes the models trained using our distillation objectives.
}
\label{table:ablation}
\end{table*}
\paragraph{Image-Text Retrieval}
To explore the generalization of our \our{} framework beyond VLU, we conduct experiments on image-text retrieval.
Table~\ref{table:retrieval_results} reports the results of the Flickr30K dataset.
Our model achieves substantial speedup with competitive performance compared to the teacher model ViLT.
The student model \our{} even outperforms the ViLT teacher in image retrieval.
Furthermore, removal of cross-modal attention distillation substantially harms performance on all metrics, showing that cross-modal attention distillation is also effective in image-text retrieval.
\begin{table}[!t]
	\centering	
	\small
    \resizebox{\linewidth}{!}{
	\begin{tabular}	{l |  c  c  c  c  c  }
	\toprule	 	
	 \multirow{2}{*}{\textbf{Methods}} & \multicolumn{2}{c}{\textbf{NLVR2}} & \multicolumn{2}{c}{\textbf{SNLI-VE}} & \textbf{VQA} \\
	  & dev & test-P & val & test & test-dev \\
	  \midrule
	    \our{} & \textbf{67.6} & \textbf{68.2} & \textbf{75.4} & \textbf{75.0} & \textbf{67.1} \\
	    \midrule
	   \quad - Soft Label & 66.8 & 67.9 & 74.2 & 74.7 & 66.9 \\
	   \quad - Cross-model Attn & 50.9 & 51.1 & 73.6 & 73.5 & 66.5 \\
	  	\quad\quad + Hidden States & 56.5 & 56.0 & 71.5 & 71.3 & 62.6 \\
        \quad\quad + Uni-Modal Attn & 64.8 & 65.6 & 74.8 & 74.8 & 66.6 \\
	  	\quad\quad + Whole Attn & 64.7 & 66.0 & 74.9 & 74.7 & 66.6  \\
		\bottomrule
	\end{tabular}
 	}
	\caption{
		Effects of using different knowledge distillation objectives.
		``Attn'' is short for attention distributions.
		``Whole Attn'' is the combination of ``Uni-modal Attn'' and ``Cross-modal Attn''.
	}
	\label{table:different_distill_strategies}
	
\end{table}		

\subsection{Analysis}
\label{section:discussion}

\paragraph{Effects of Distillation in Training Stages}
We investigate the effect of applying our proposed distillation in the pre-training and fine-tuning stages.
We compare with the baseline trained without distillation objectives, as the standard training.

Table~\ref{table:ablation} shows the evaluation results.
Without the pre-training initialization, the models are directly initialized by the weights of pretrained ViLT.
We can observe that without the pre-training stage, the performance significantly drops arcoss tasks.
Under this setting, without the proposed distillation training, the model obtains a chance-level performance on NLVR2, similar to previous work~\cite{vilt,how_can_clip}.
But our distillation method substantially improves the results, reducing the gap from $-12.40$ to $-3.94$.
This indicates that our distillation method is better than standard training, even without pre-training.
Furthermore, in the pre-training stage, the model still benefited from the distillation objectives compared with the standard training.
Another interesting observation is that fine-tuning distillation brings more gains compared to pre-training distillation.
This suggests that it is crucial for dual-encoder models to learn more ability of the task-specific cross-modal interactions.
Overall, performing our proposed method in both pre-training and fine-tuning stages delivers the best performance across VLU tasks.

\paragraph{Effects of Different Distilled Knowledge}
We investigate the effects of different knowledge used in our framework.
The compared ablations include training without soft label distillation ($-$ soft label) or cross-modal attention distillation ($-$ cross-modal attn).
For more clarity, the dual-encoder student models are directly initialized by the pretrained ViLT and fine-tuned with varying distillation objectives.

Table~\ref{table:different_distill_strategies} illustrates the results of the VLU tasks.
We find that both distillation objectives contribute to the success of the \our{} framework, while the proposed cross-modal attention distillation is more critical than soft label distillation.
The student only trained with the soft label distillation objective only achieves random performance on NLVR2.
We further incorporate other intermediate representations of the teacher model, except for cross-modal attention.
We observe that using attention distributions brings more gains across three tasks compared to the hidden states.
Furthermore, we also explore which part of the attention distribution is more critical, \textit{cross-modal attention} or \textit{uni-modal attention}.
As shown in Table~\ref{table:different_distill_strategies}, mimicking the teacher's cross-modal attention distributions achieves more improvements.
Furthermore, we find that only the use of cross-modal attention distributions performs better than using the whole attention distributions (cross-modal $+$ uni-modal).
These observations validate our motivation that cross-modal interactions are more crucial for VLU tasks.

\begin{table}[!t]
    \resizebox{\linewidth}{!}{
	\begin{tabular}	{l |  c  c  c  c  c  }
	\toprule	 	
	 \multirow{2}{*}{\textbf{Methods}} & \multicolumn{2}{c}{\textbf{NLVR2}} & \multicolumn{2}{c}{\textbf{SNLI-VE}} & \textbf{VQA} \\
	  & dev & test-P & val & test & test-dev \\
	  \midrule
 Last Layer (Ours) & \textbf{67.6} & \textbf{68.2} & \textbf{75.4} & \textbf{75.0} & \textbf{67.1}\\
	    Top-Layers Layerwise & 66.0 & 67.3 & 75.2 & 74.8 & 66.8\\
	    Bottom-Layers Layerwise & 63.5 & 63.0 & 75.1 & 74.7 & 66.6\\
        All-Layers Layerwise & 67.0 & 67.4 & 75.2 & 74.8 & 66.8 \\
		\bottomrule
	\end{tabular}
 	}
    \caption{
		Effects of different layer mapping strategies for our distillation method.
	}
	\label{table:layer_mapping_function}
	
\end{table}		
\paragraph{Effects of Different Layer Mapping Strategies for Distillation.}
To validate the effectiveness of distilling the knowledge of the teacher last layer for the student, we compare it with the layer-wise mapping strategy, including all layers, the upper part of layers, and the bottom part of layers.
As shown in Table~\ref{table:layer_mapping_function},
last-layer strategy obtains better results.
Furthermore, our strategy requires less computation than the layerwise methods.
Thus, distillation in the last layer is more practical.

\begin{figure}[t]
    \begin{center}
    \includegraphics[trim={0 9.0 0 0.5}, clip, width=\linewidth]{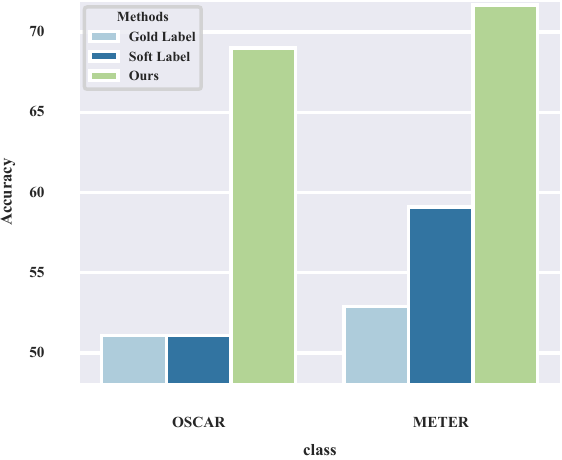}
    \caption{ 
        NLVR2 results of models initialized by OSCAR or METER.
        ``Ours'' means that besides the soft label distillation, the dual-encoder model also trained with our proposed cross-modal attention distillation.
    }
    \label{fig:other_models_results}
    \end{center}
\end{figure}

\paragraph{Effects on other VLP models.}
To evaluate the generalization of \our{}, in addition to ViLT~\cite{vilt}, we also conduct experiments on NLVR2 with two other fusion-encoder vision-language pretrained models, OSCAR~\cite{oscar} and METER~\cite{meter}.
The architectures of two models are different from ViLT:
Oscar applies the pretrained object detector to extract visual features.
METER uses the visual encoder of CLIP to encode image patches and also applies RoBERTa~\citep{roberta} as textual encoder. The visual and textual representations are then fused by a multi-layer Transformer network.

We directly adopt the pretrained models to initialize the dual-encoder model and take the fine-tuned models as the teacher.
We compare our method with the baselines supervised only by the gold labels of the original dataset and the soft label from the teacher model.
Figure~\ref{fig:other_models_results} illustrates the performance of the methods.
For the models initialized by OSCAR and METER, we observe that only fine-tuning with gold labels or soft labels can not benefit the performance of the dual-encoder model on the complex visual reasoning task NLVR2.
On the contrary, our method can improve performance by a large margin on both models, which is consistent with the results on ViLT.
Our proposed cross-modal attention distillation shows effectiveness on different architectures of vision-language models. 
\section{Conclusion}
On vision-language understanding tasks, fusion-encoder models obtain superior performance while sacrifice efficiency.
In contrast, dual-encoder models have the advantage of efficiency, but previous models are insufficient to handle complex vision-language understanding.
In this work, to obtain the efficiency-performance trade-off, we propose \our{}, a knowledge distillation framework for dual-encoder models to improve their performance on VLU tasks while retaining their efficiency.
The key of \our{} is that we employ the cross-modal attention of a fusion encoder model as fine-grained supervision to guide the dual-encoder model for learning complex cross-modal interactions.
Experimental results on several vision-language understanding tasks show that our \our{} achieves competitive performance with a four-time speedup over the fusion-encoder teacher model.
Further analyses verify that distillation with cross-modal attention is critical for dual-encoder models.

\section*{Limitations}
\our{} is pretrained on the publicly accessible resources consisting of $4$M image-text pairs.
We do not have enough computational resources to explore the situation with larger data, as used in CLIP~\citep{clip}.
It is also interesting to combine our approach with other model acceleration methods summarized in~\citet{green_survey} to further toward Green AI~\citep{green_ai}.

\bibliography{anthology,custom}
\bibliographystyle{acl_natbib}

\appendix
\section{Details of Hyperparameters}
\label{sec:app:hyperparam}
For pre-training, visual and textual encoders of \our{} are initialized by the weights of the teacher model. 
We use Adam~\cite{adam} with $\beta_1=0.9, \beta_2=0.999$ for optimization.
The learning rate is set to 1e-4, with the warm-up ration of $0.1$, and linear decay.
The weight decay is set to $0.01$.
For the ITM task, we replace the matched image with the probability of 0.5 to construct negative examples following previous work~\cite{uniter, oscar, vilt}, 
For the MLM task, we use 15\% masking probability as in BERT~\cite{bert}.
For the downstream fine-tuning, we follow most of the hyperparameters in~\citet{vilt}.
We fine-tune the model for $10$ epochs with a batch size of $256$ for VQA and SNLI-VE.
For NLVR2, we train the model for $20$ epochs with a batch size of $128$.
For Flickr30k, the model is trained for $20$ epochs with a batch size of $1024$. 
We apply RandAugment~\cite{randaugment} without color inversion and cutout.

\end{document}